\documentclass[10pt,journal,compsoc]{IEEEtran}

\ifCLASSOPTIONcompsoc
  \usepackage[nocompress]{cite}
\else
  \usepackage{cite}
\fi

\usepackage{acronym}
\usepackage{cite}
\usepackage{graphicx}
\usepackage{url}
\usepackage{subcaption}
\usepackage{tikz}
\usepackage{pgfplots}

\acrodef{1D}{one-dimensional}
\acrodef{2D}{two-dimensional}
\acrodef{2.5D}{two-and-a-half-dimensional}
\acrodef{3D}{three-dimensional}
\acrodef{CNN}{Convolutional Neural Network}
\acrodef{ConvLSTM}{Convolutional LSTM}
\acrodef{DoF}{Degrees of Freedom}
\acrodef{ECC}{Edge-Conditioned Convolution}
\acrodef{ELU}{Exponential Linear Unit}
\acrodef{k-NN}{k-Nearest Neighbors}
\acrodef{LSTM}{Long Short-Term Memory Network}
\acrodef{MLP}{Multi Layer Perceptron}
\acrodef{MNIST}{Modified National Institute of Standards and Technology}
\acrodef{NELL}{Never-Ending Language Learning}
\acrodef{GCN}{Graph Convolutional Network}
\acrodef{GNN}{Graph Neural Network}
\acrodef{GPU}{Graphics Processing Unit}
\acrodef{ReLU}{Rectified Linear Unit}
\acrodef{RF}{Random Forest}
\acrodef{ROS}{Robot Operating System}
\acrodef{SVM}{Support Vector Machine}

\begin{document}
\title{TactileGCN: A Graph Convolutional Network for Predicting Grasp Stability with Tactile Sensors}
\author{A.~Garcia-Garcia*,
        B.S.~Zapata-Impata*,
        S.~Orts-Escolano,
        P.~Gil,
				J.~Garcia-Rodriguez%
\IEEEcompsocitemizethanks{\IEEEcompsocthanksitem 
A.~Garcia-Garcia, S.~Orts-Escolano, and J.~Garcia-Rodriguez are with the 3D Perception Lab, University of Alicante, Spain. B.S.~Zapata-Impata and P.~Gil are with the Automatics, Robotics and Artificial Vision Lab, University of Alicante, Spain.\protect\\
E-mail: agarcia@dtic.ua.es, brayan.impata@ua.es, sorts@ua.es, pablo.gil@ua.es, jgarcia@dtic.ua.es
}% <-this % stops an unwanted space
\thanks{*Authors contributed equally.}}
\markboth{}%
{Shell \MakeLowercase{\textit{et al.}}: Bare Demo of IEEEtran.cls for Computer Society Journals}
\IEEEtitleabstractindextext{%
\begin{abstract}
Tactile sensors provide useful contact data during the interaction with an object which can be used to accurately learn to determine the stability of a grasp. Most of the works in the literature represented tactile readings as plain feature vectors or matrix-like tactile images, using them to train machine learning models. In this work, we explore an alternative way of exploiting tactile information to predict grasp stability by leveraging graph-like representations of tactile data, which preserve the actual spatial arrangement of the sensor's taxels and their locality. In experimentation, we trained a \acl{GNN} to binary classify grasps as stable or slippery ones. To train such network and prove its predictive capabilities for the problem at hand, we captured a novel dataset of $\sim5000$ three-fingered grasps across $41$ objects for training and $~1000$ grasps with $10$ unknown objects for testing. Our experiments prove that this novel approach can be effectively used to predict grasp stability.
\end{abstract}
\begin{IEEEkeywords}
Graph Neural Networks, Tactile Sensors, Grasping, Grasp Stability, Deep Learning, Robotics
\end{IEEEkeywords}}
\maketitle
\IEEEdisplaynontitleabstractindextext
% \fi
\IEEEpeerreviewmaketitle

\IEEEraisesectionheading{\section{Introduction}\label{sec:introduction}}
\IEEEPARstart{W}{hen} we humans grasp objects, we know whether the grip is stable or not before lifting the object. It is not necessary to raise our hands in order to check such state of the grasp. Using our tactile sense, along with our vision and other senses, we can accurately predict the stability of the grasp. This skill is desirable for any robotic manipulator since it favors the early detection of grasp failures so the robot can react in consequence: for example, a re-stocking robot working in a store would recognize when an object could slip from its hand and, therefore, avoid breaking it.

\begin{figure}[t]
	\centering
	\includegraphics[width = 0.49\textwidth, clip = false, trim = 0 0 0 0]{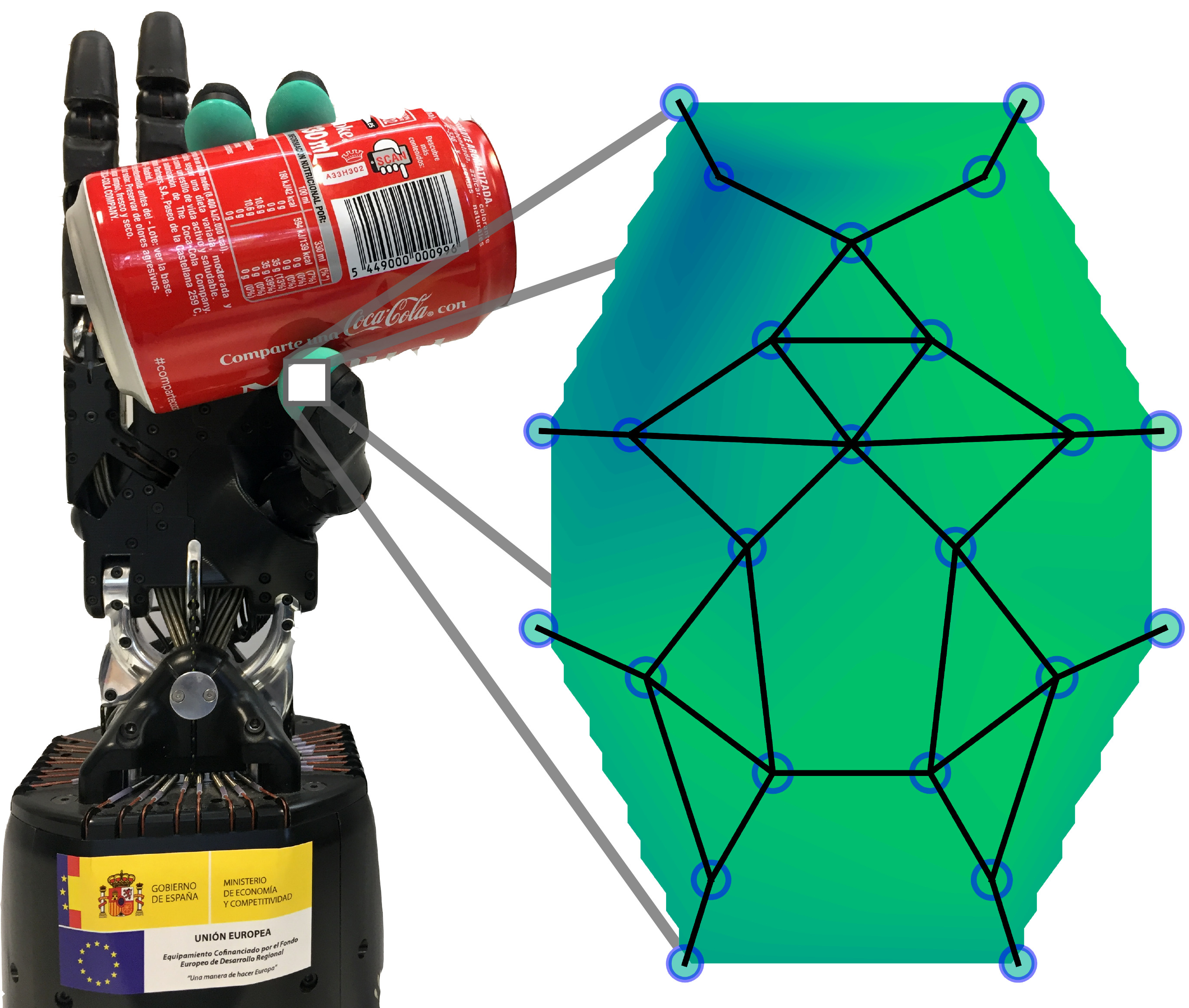}
	\caption{In this work, we use a Shadow Dexterous hand equipped with three BioTac SP tactile sensors whose readings are transformed into graph representations. Those graphs are then fed as input to a \ac{GNN} to learn to predict grasp stability.}
	\label{fig:shadow-coke}
\end{figure}

The problem of predicting the stability of a grasp is a task under research in the field of robotic grasping. In order to approach a solution to it, tactile sensors are being used as the main source of data since they provide valuable information (e.g. temperature, pressure) about the acting forces during the interaction of the robotic hand with the objects \cite{Kappassov2015}. As for the stability prediction, two states are usually distinguished: stable, meaning that the object is firmly grasped; or slippery, meaning that the object could slide from the hand.

Previous works found in the literature approach this problem following the next methodology: grasp the object, read the tactile sensors equipped in the fingers and/or palm of the hand, calculate custom features that try to characterize these two stability states and learn them in order to make future predictions \cite{Li2014,Dang2014,Su2015,Veiga2015}. These proposals treat the tactile readings as classic signals: they pre-process them as if they were arrays, calculate features and learn their characteristics using probabilistic methods. As a consequence, their performance highly depends on the selected characteristics. Moreover, the spatial distribution inherent to the tactile sensor is lost due to the fact of squeezing the data into a one dimensional array.

In this work, we propose the use of \acp{GNN} for predicting grasp stability. Since these are deep learning models, there is no need to hand-engineer features because the algorithm is designed for learning them by itself. Moreover, graphs can reflect more accurately the real distribution of the electrodes in the sensor as well as their spatial relationships, which should be of great value for learning tactile features. The main contributions of this work can be summarized as follows:

\begin{itemize}
	\item We process tactile readings using a novel perspective: instead of considering them as 1D arrays or 2D images, we build a 3D graph connecting the multiple sensing points (taxels) of the tactile sensor.

	\item We introduce a novel way of processing such information using \acfp{GNN}.
	
	\item We quantitatively check the performance of this new methodology in the real world using a set of tactile sensors installed in a robotic hand, seen in Figure \ref{fig:shadow-coke}.

	\item We release an extension that effectively doubles the size of an already existing dataset \cite{Zapata2018} for grasp stability prediction and includes a whole new split for testing.
\end{itemize}

This paper is organized as follows. Section \ref{sec:related_works} reviews the state of the art of grasp stability prediction using tactile sensors and \acp{GNN}. Section \ref{sec:proposal} describes our system from the tactile graphs generation process to the network architecture. Section \ref{sec:experimentation} contains the methodology and data used to validate our proposal, as well as quantitative results to support our claims. Section \ref{sec:conclusion} summarizes our findings and contributions. At last, Section \ref{sec:limitations} states the main limitations of this work and draws some future research lines.

\section{Related Works}
\label{sec:related_works}

In this section, we review the state of the art of the two main fields related to our work. On the one hand, we describe previous approaches for predicting grasp stability. On the other hand, we explain the most recent and relevant advances in neural networks for graph processing.

\subsection{Grasp Stability Prediction}
\label{sec:grasp-stability-prediction}

In the last years, deep learning models are being applied to the problem of grasp stability prediction using tactile sensors as input. Meier \emph{et al.} \cite{Meier2016a} processed tactile readings using Fourier-related transforms and the resulting vectors were vertically stacked in order to create a matrix. Then, a \ac{CNN} trained with these matrices learnt to predict stability. Although this approach used modern machine learning models, it still had to hand-engineer features.

In contrast, Cockbum \emph{et al.} \cite{Cockbum2017} proposed to use autoencoders to autonomously calculate the relevant characteristics for the task. Afterwards, a dictionary of basis features was built using a sparse encoding algorithm. Finally, the authors trained a \ac{SVM} in order to predict grasp stability using the dictionary. Similarly, Kwiatkowski \emph{et al.} \cite{Kwiatkowski2017} built a composite image by placing the readings of two matrix-like sensors side by side. Then, they used this tactile image as input for a \ac{CNN} along with the proprioceptive data from the robot. As a result, the proposed method calculated by itself the features needed for predicting grasp stability.

A more recent trend suggests the interpretation of tactile sensors as images in order to exploit the potential of \ac{CNN} as feature learners. In some cases, vision-based sensors are used for this purpose. Calandra \emph{et al.} \cite{Calandra2017} used a tactile sensor that contained an internal camera, which recorded the deformation of the gel inside of the sensor throughout its contact with a surface. Then, the recorded tactile images were learnt using a \ac{CNN} in order to predict the grasp outcome. In some other cases, the tactile sensor is not naturally arranged in an array or it does not contain a camera, so a pre-processing is necessary in order to get a tactile image. For example, Zapata-Impata \emph{et al.} \cite{Zapata2018} studied how the readings from a non-matrix like sensor should be arranged in a matrix in order to train a \ac{CNN} for grasp stability prediction. Although such approach showed promising results, the spatial distribution of the real sensor was not accurately reflected because it reduced the \acs{3D} locations of the taxels into \acs{2D} coordinates of a tactile image.

Recently, \acp{CNN} are being combined with \acp{LSTM} for grasp stability prediction. Li \emph{et al.} \cite{Li2018} in their work learnt visual features from a camera-based tactile sensor, similar to the one used by Calandra \emph{et al.} \cite{Calandra2017}, and an external camera pointing to the scene. These features were calculated using a pre-trained \ac{CNN}. Then, both cameras features were concatenated and passed in time sequences to a \ac{LSTM}, which was in charge of detecting slippage. Similarly, Zhang \emph{et al.} \cite{Zhang2018} used another camera-based tactile sensor for grasp stability detection but in this work the authors trained a \ac{ConvLSTM} and they only passed the sensor images to the network.

% BRAYAN: Como nos pasamos ya de 8 paginas, crees que podrias resumir esta subseccion? la parte de estabilidad ha quedado en 1 columna, esta tiene 2. Igual podemos ahorrar en espacio.

\subsection{Graph Neural Networks}
\label{sec:gcns}

Lately, \acp{GNN} have emerged as a solid alternative to process irregular data which can be structured as graphs. Their original focus was tasks whose data can be expressed as graphs holding locality, stationarity, and composionality principles in general. In the literature, various works have successfully made use of this kind of architecture to deal with unstructured \acs{3D} representations mainly in classification tasks. Most of them  have proposed extensions to the well-known \ac{CNN} architecture to process graph-structured data. That generalization is not trivial since various problems must be addressed when applying convolution filters in domains in which there is no regular structure. In that regard, there are two dominant ways to convolve a graph signal with a learned filter: spatial or spectral.

Spectral methods are characterized by providing a spectral graph theoretical formulation of CNNs on graphs using Graph Signal Processing (GSP) theory \cite{Shuman2013}. The fundamentals of this kind of methods rely on decomposing the graph Laplacian to form a Fourier basis via an eigendecomposition of the graph matrix, i.e., a spectral decomposition. By doing that, a convolution in the graph domain can be expressed as a multiplication in the spectral one. This kind of methods usually faces three challenges: the design of compactly supported filters, the definition of parameter sharing schemes among different graphs, and the aggregation of multi-scale information. Arguably, the most common and limiting drawback is the first challenge: filters are not directly transferable to different graphs. Since filters are learned in the context of the spectrum of the graph Laplacian, a global graph structure must be assumed. In other words, only the signals on the vertices may change, the structure of the graph must remain the same.

Spatial methods constitute the straightforward generalization of convolutions to graph, just by sliding a filter on the vertices as a traditional CNN does with any other structured data representation. Despite its simplicity, the direct application of the definition of a convolution to graphs poses two difficulties: the definition of neighborhoods, and the ordering of the nodes to form receptive fields. Because of that, one common problem of spatial methods is the difficulty to generate a weight sharing schema across graph locations due to the fact that local neighborhoods can be completely different, i.e., the number of nodes adjacent to another one varies and there is no well-defined ordering for them.

Here we briefly review the most relevant \acp{GNN} that have been successfully applied to similar problems to the one at hand.

The pioneer spectral formulation of a CNN to operate over irregular domains modeled as graphs was introduced by Bruna et al. \cite{Bruna2013}. In that work, they exploited the global structure of the graph with the spectrum of its graph-Laplacian to extend the convolution operator. This method was applied to hand-written digit classification using the \ac{MNIST} dataset.

Defferrard \emph{et al.} \cite{Defferrard2016} proposed strictly localized filters, which are provable to be localized in a ball of a certain radius, i.e., hops from a specific vertex. That enhancement has some other collateral effects such as improved computational complexity for the filters (linear w.r.t. the support’s size and the number of edges). They also introduced an efficient  pooling strategy based on a rearrangement of the vertices as a binary tree. Their approach, namely \emph{Chebyshev Spectral Graph Convolutional Operator} or just \emph{ChebConv}, was successfully applied and performed similarly to classical \acp{CNN} in digits classification problems such as \acs{MNIST}.

Kipf and Welling \cite{Kipf2016} introduced a set of simplifications to Bruna's \cite{Bruna2013} and Defferrard's \cite{Defferrard2016} formulations to improve performance and scalability in large-scale networks. They proved the efficacy of their work on transductive node classification on very large scale networks for various problems such as semi-supervised document classification in citation networks (CiteSeer, Cora and PubMed datasets) and semi-supervised entity classification in a knowledge graph (\acs{NELL} dataset). As its main feature, their \emph{GCNConv} operator takes advantage of fast localized first-order features to achieve linear scaling in the number of graph edges.

Simonovsky and Komodakis \cite{Simonovsky2017}, inspired by the idea from Jia \emph{et al.} [3] about dynamic filter networks, took a similar approach for solving the weight sharing problem suffered by spatial methods. They introduced \acp{ECC} in which filter weights are conditioned on edge features and generated by a generator network. That generator, usually implemented as a \ac{MLP}, outputs specific weights for each edge in the neighborhood. That method was successfully tested on point cloud classification problems (Sydney urban objects and ModelNet), a standard graph classification benchmarks, and also on \acs{MNIST}.

Velickovic \emph{et al.} \cite{Velickovic2017} introduced a \emph{Graph Attention} operator, namely \emph{GATConv}, that leverages masked self-attentional layers to compute the hidden representations of each node in the graph, by attending over its neighbors, following a self-attention strategy. This approach addressed many of the key challenges of spectral-based methods and achieved or surpassed state of the art methods in the aforementioned citation network datasets as well as protein interaction ones.

Fey \emph{et al.} \cite{Fey2018} proposed the \emph{Spline-based Convolutional Operator}, a continuous and spatial kernel that leverages B-spline bases's properties to efficiently filter graph data of arbitrary dimensionality. They prove this method to be successful in digit image graph classification problems using \acs{MNIST} and graph node classification using the Cora dataset.

%\subsection{Our Proposal in Context}

Following this success in those similar domains, we intend to use a \acs{GNN} to process tactile sensor readings and predict grasp stability. By doing so, we expect that such architecture is able to better capture the spatial locality and relationships of the tactile sensor readings expressed as graphs instead of other non-spatial (\acs{1D} arrays) or discrete (images) representations.

\section{Proposal}
\label{sec:proposal}

In this section, we describe our full approach for predicting grasp stability using tactile sensors. The whole pipeline comprises three main components:

\begin{enumerate}
    \item A robotic setup which consists of a Shadow hand and BioTac Sp sensors, all operated by \ac{ROS}.
    \item A tactile graph generator which takes the sensor readings and generates a proper graph representation for the network.
    \item A \ac{GNN} architecture to process such graphs and predict graph stability.
\end{enumerate}

%The remainder of this section will describe each of those components in depth.

\subsection{Robotic Set Up}
\label{sec:rpobotic-set-up}

\begin{figure}[!htb]
	\centering
	\includegraphics[width = 0.26\textwidth, clip = true, trim = 0 75 0 20]{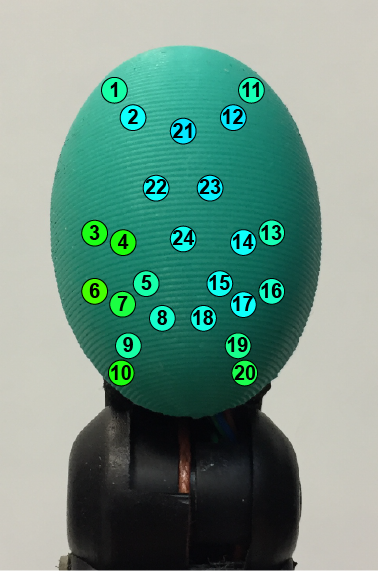}
	\caption{BioTac SP tactile sensor with its 24 electrodes approximated position.}
	\label{fig:biotac-sensor}
\end{figure}

In this work, we use the BioTac SP tactile sensors developed by Syntouch \cite{Syntouch2018}. The sensor provides three different sensory modalities: force, pressure, and temperature. In more detail, this biomimetic sensor counts with $24$ electrodes, also named taxels, integrated in just a single phalanx. These electrodes record signals from four emitters in the internal core of the sensor and, therefore, they measure the impedance in the fluid located between the internal core and the external elastic skin of the sensor. The fluid is displaced when the sensor makes contact with a surface, affecting that impedance read by the electrodes. Thus, the sensor can approximate how much pressure is being experienced at each electrode. In addition, the sensor features a hydro-acoustic pressure sensor in order to estimate a general pressure value and it also counts with a thermistor, which is used to detect vibrations and heat flows. The sensor is presented in Figure \ref{fig:biotac-sensor}.

For our work, we use a setup of three BioTac SP sensors in the tip of the index, middle finger, and thumb of a Shadow Dexterous robotic hand developed by the Shadow Robot Company \cite{ShadowRobotCompany2018}. The Shadow hand is an anthropomorphic hand with five fingers and $20$ \ac{DoF} in total. Those features allow the robot to reach a wide range of configurations that are comparable to those of a human hand. Its integration with the BioTac SP sensors is seamless since the sensor readings can be directly obtained using the \ac{ROS} \cite{Quigley2009} framework, in which the Shadow hand works.

\subsection{Tactile Graphs}
\label{sec:tactile-graphs}

In order to feed our \acl{GNN}, we expressed the aforementioned sensor readings in a novel graph representation, namely tactile graphs. Such graphs are triplet $G = (N, E, Y)$ where $N$ is a set of $24$ nodes ${n_0, ..., n_{23}}$ (one for each electrode or taxel in the sensor), $E$ is a set of ordered pair of vertices called edges, and $Y$ is the label or class of the graph (in our case, stable or unstable).

Each node $n$ in the graph $G$ represents a taxel and as such, they are characterized by a \acs{3D} position $p_n = (x_n, y_n, z_n)$ and a feature vector $f_n = (f_{n_0}, ..., f_{n_F})$ of arbitrary length $F$.

\begin{figure}[!htb]
	\centering
    \includegraphics[width=0.8\linewidth]{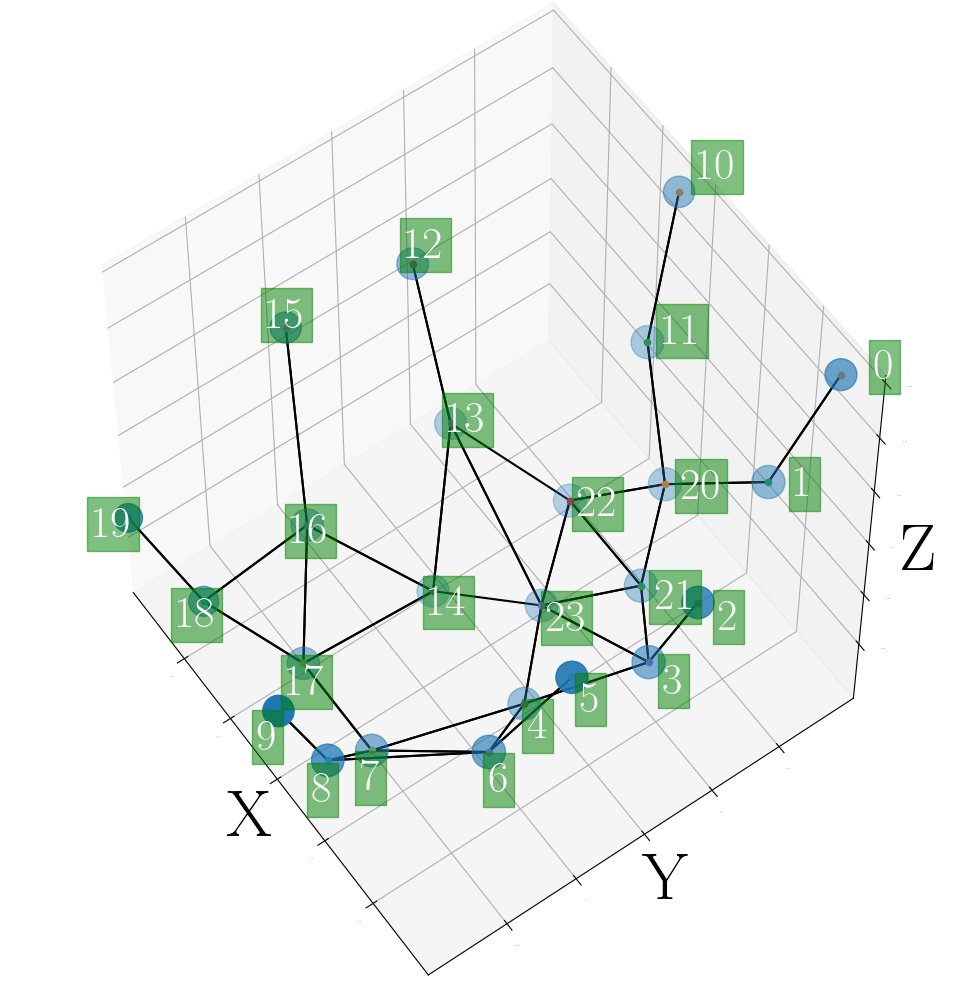}
	\caption{3D visualization of the tactile graph layout using the accurate spatial arrangement from the actual BioTacSp sensor. Graph edges correspond to the manually defined connections.}
	\label{fig:graph_3d}
\end{figure}

Node positions $p_n$ are accurately mapped to the physical taxel $(X, Y, Z)$ coordinates within the sensor. Such positions are specified in Table \ref{table:taxel_coordinates}. Edges or connections are generated following two different approaches: manual or \ac{k-NN}. For the first approach, we manually specified undirected connections following proximity and symmetry criteria. For the second one, we generated directed edges towards each \acl{k-NN} for each node. Figure \ref{fig:graph_3d} shows a \acs{3D} graph representation of a tactile graph.

% ALBERT: Lo pongo como unidades en inch. ¿el orden es correcto? XYZ? Me pongo con lo de la Figura 3.
% BRAYAN: Bueno, el orden es el que les queramos dar.

\begin{table}[!htb]
    \centering
    \caption{Taxel positions inside the BioTacSp sensors expressed in cartesian coordinates $(X,Y,Z)$ in inches.}
    \begin{tabular}{lccc}
        \hline
        \textbf{Taxel} & \textbf{X (inches)} & \textbf{Y (inches)} & \textbf{Z (inches)}       \\
        \hline
        1     & 0.386434851  & -0.108966104 & 0.156871012  \\
        2     & 0.318945051  & -0.205042252 & 0.120706090  \\
        3     & 0.087372680  & -0.128562247 & 0.281981384  \\
        4     & 0.083895199  & -0.235924865 & 0.201566857  \\
        5     & -0.018624877 & -0.300117050 & 0.094918748  \\
        6     & -0.091886816 & -0.120436080 & 0.284956139  \\
        7     & -0.136659500 & -0.237549685 & 0.187122746  \\
        8     & -0.223451775 & -0.270674659 & 0.071536904  \\
        9     & -0.320752549 & -0.199498368 & 0.127771244  \\
        10    & -0.396931929 & -0.100043884 & 0.151565706  \\
        11    & 0.386434851  & -0.108966104 & -0.156871012 \\
        12    & 0.318945051  & -0.205042252 & -0.120706090 \\
        13    & 0.087372680  & -0.128562247 & -0.281981384 \\
        14    & 0.083895199  & -0.235924865 & -0.201566857 \\
        15    & -0.018624877 & -0.300117050 & -0.094918748 \\
        16    & -0.091886816 & -0.120436080 & -0.284956139 \\
        17    & -0.136659500 & -0.237549685 & -0.187122746 \\
        18    & -0.223451775 & -0.270674659 & -0.071536904 \\
        19    & -0.320752549 & -0.199498368 & -0.127771244 \\
        20    & -0.396931929 & -0.100043884 & -0.151565706 \\
        21    & 0.258753050  & -0.252337663 & 0.000000000  \\
        22    & 0.170153841  & -0.274427927 & 0.072909607  \\
        23    & 0.170153841  & -0.274427927 & -0.072909607 \\
        24    & 0.075325086  & -0.298071391 & 0.000000000\\
        \hline
    \end{tabular}
    \label{table:taxel_coordinates}
\end{table}

% BRAYAN: Se puede sacar otra vez la figura 3? Por sumarle 1 a la numeracion y asi hacerla coherente con el resto (figura 2 y tabla 1).

\begin{figure}[!htb]
    \centering
    \includegraphics[width=\linewidth, clip, trim={90 190 80 0}]{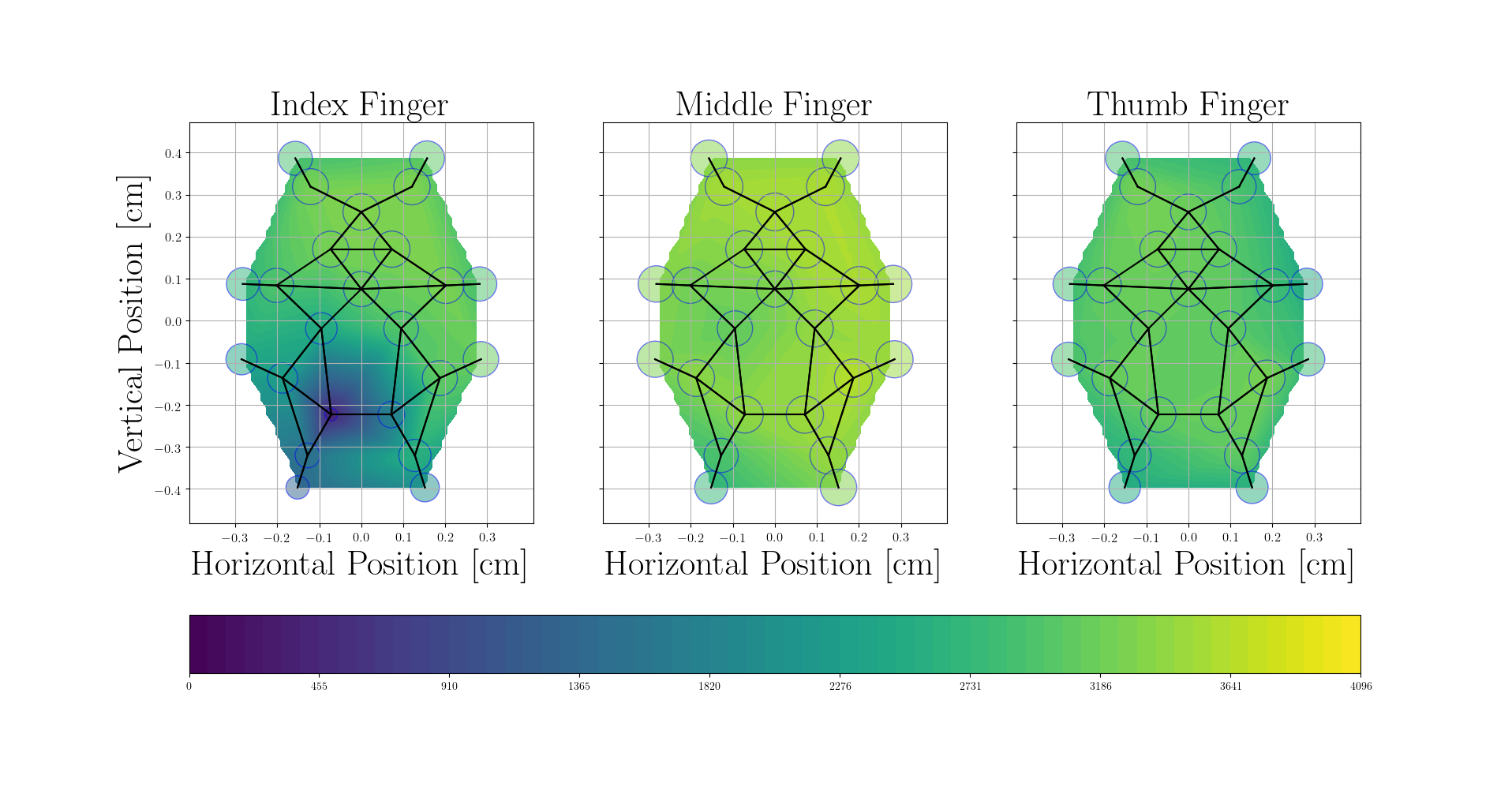}
    \bigskip
    \includegraphics[width=\linewidth, clip, trim={90 190 80 0}]{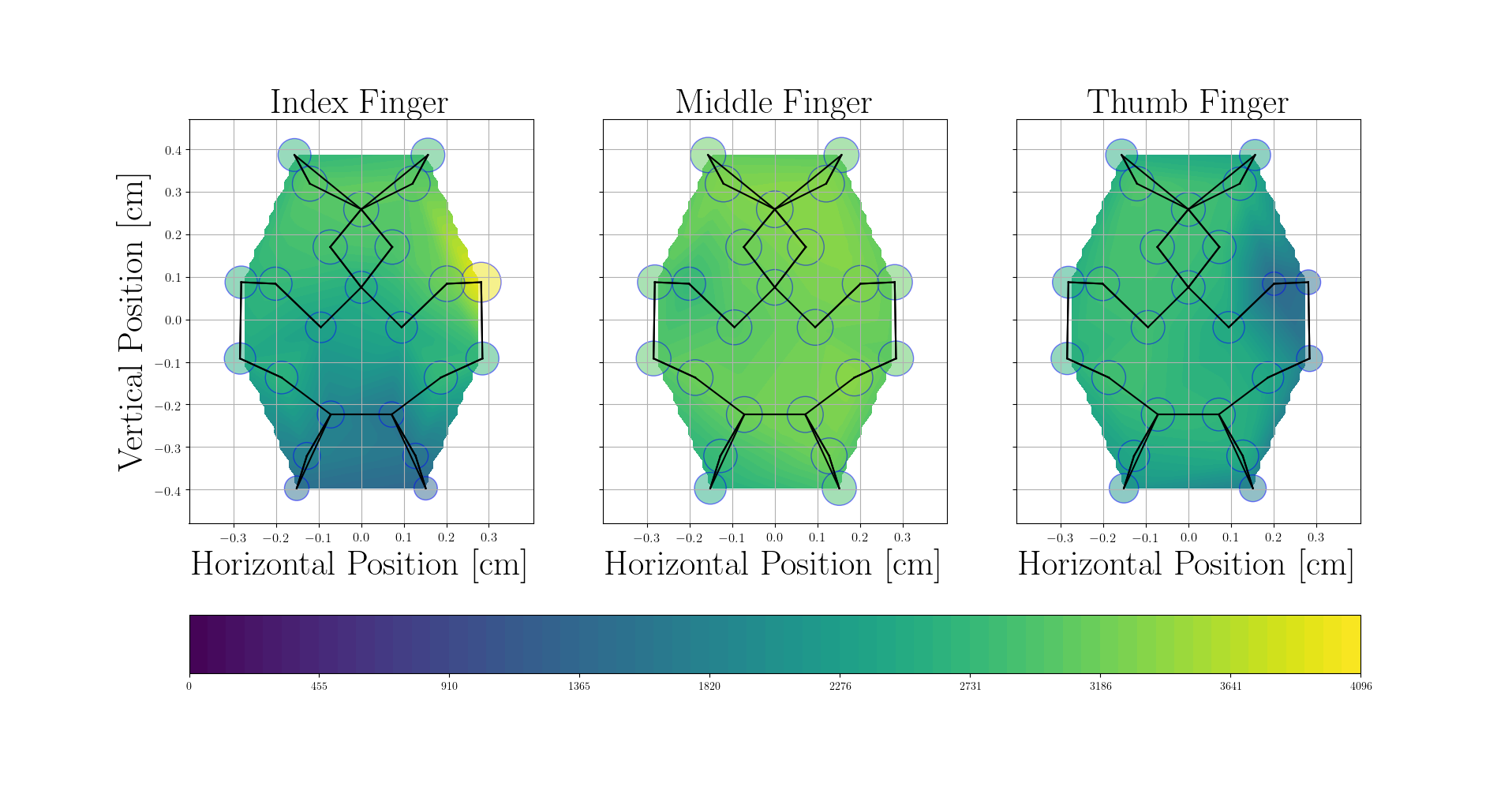}
    \bigskip
    \includegraphics[width=\linewidth, clip, trim={90 190 80 0}]{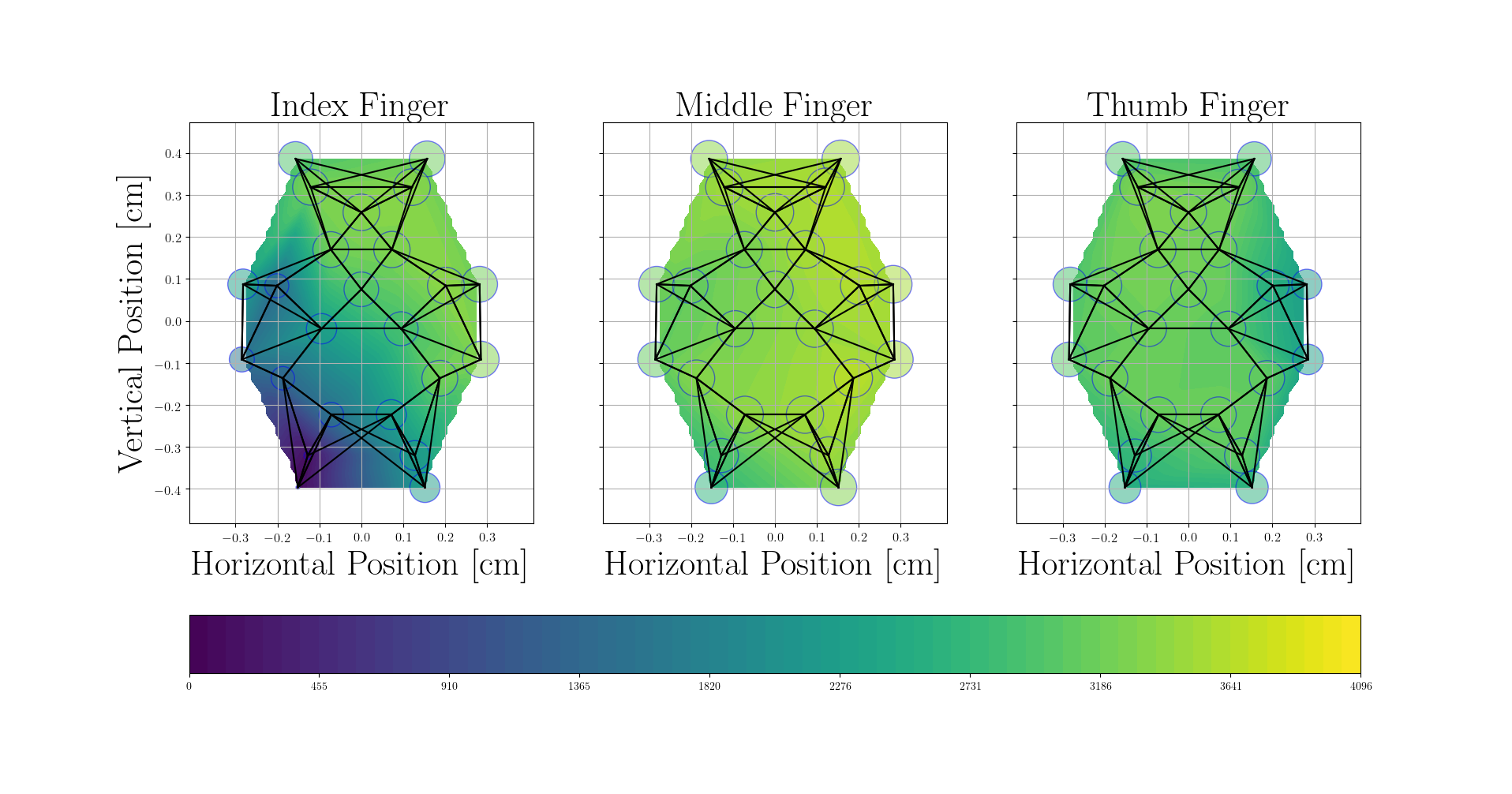}
    \bigskip
    \includegraphics[width=\linewidth, clip, trim={90 0 80 0}]{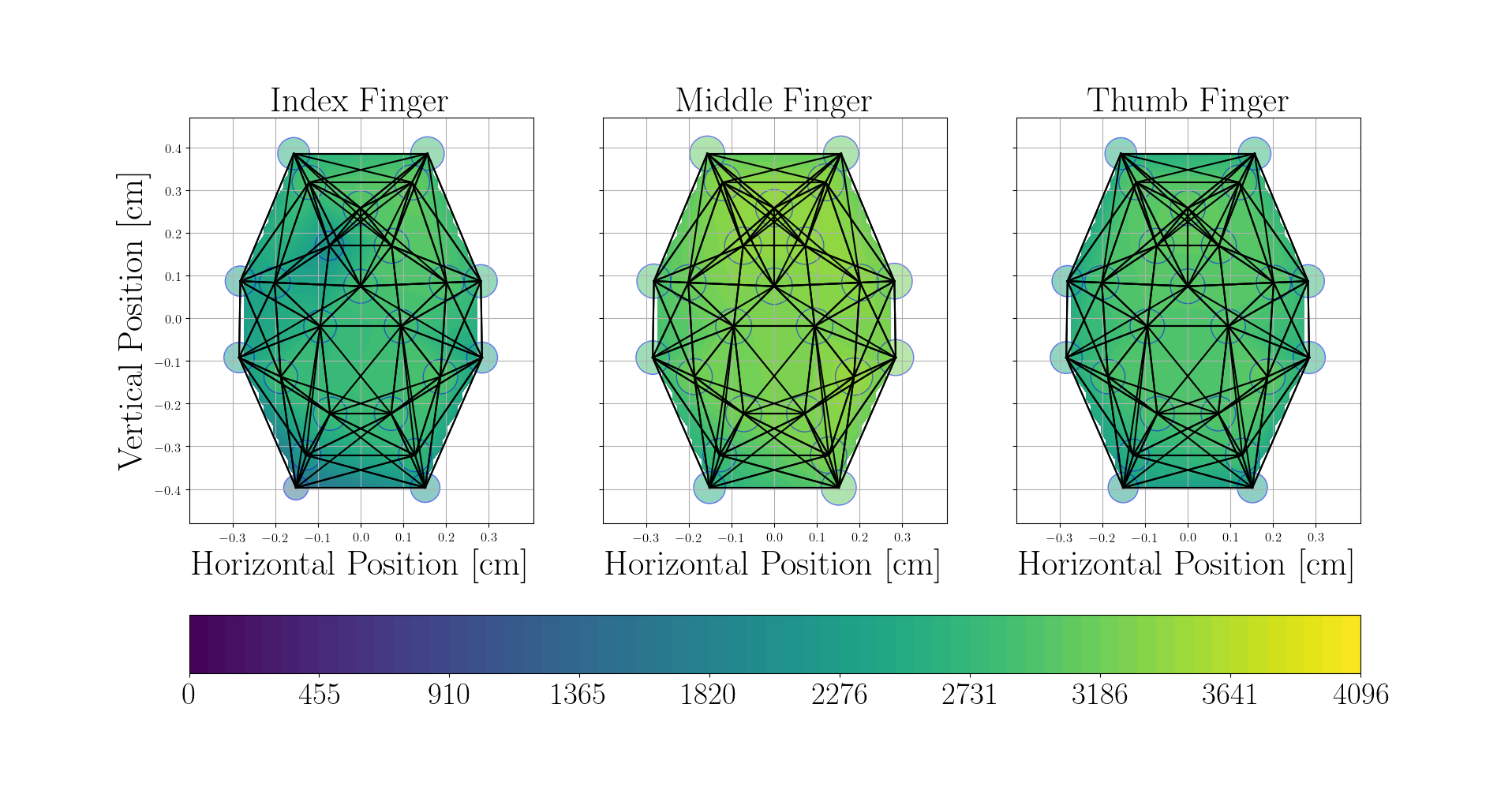}
	\caption{From top to bottom, undirected tactile graphs generated with various \ac{k-NN} configurations ($k=0$ (manually defined edges), $k=2$, $k=4$, and $k=8$). The three features $f_{n_0}$, $f_{n_1}$, and $f_{n_2}$ are deocupled into three different plots for each tactile graph and represented as contour plots in the XY plane. Nodes or taxels are shown as blue semi-transparent circles whose size depends on the pressure on them. Undirected edges are represented by black lines. Features are color-coded in the range $[0, 4096]$.}
	\label{fig:sample_graphs}
\end{figure}

Node features $f_n$ correspond to the taxel pressure readings. In the case of the most basic tactile graph, each node has three features, i.e., the pressure reading for each finger: index $f_{n_0}$, middle $f_{n_1}$, and thumb $f_{n_2}$. Figure \ref{fig:sample_graphs} shows visualizations of the three components of the feature vector for sample graphs generated with various values of $k=0$, $k=2$, $k=4$, $k=8$.

% BRAYAN: Si alguien lee el paper por encima mirando figuras, creo que ahora mismo es un poco lioso descubrir cual es la distribucion del sensor. Que tal si juntamos la figura 2 del sensor (quitandole los nodos que dibuje encima) con esta figura 4 del grafo en 2D. 
% ALBERT: Toda la razón, déjame que le de un par de vueltas.

% \begin{figure}[!htb]
% 	\centering
% 	\includegraphics[width=0.8\linewidth]{img/index_graph_k0.png}
% 	\caption{Undirected tactile graph generated representing the node features (index finger readings $f_{n_0}$) as a contour plot. Nodes or taxels are shown as blue circles whose size depends on the pressure. Undirected edges are represented by black lines. $f_{n_0}$ is color-coded in the range $[0, 4096]$.}
% 	\label{fig:index_graph_k0}
% \end{figure}

\subsection{\acl{GNN}}
\label{sec:gnn}

Our \acf{GNN} of choice is based on the \ac{GCN} model by Kipf and Welling \cite{Kipf2016}. Such model is arguably one of the most successful, yet simple, approaches to date to generalize a well-established model such as the \ac{CNN} to arbitrarily structured graphs \cite{Bronstein2017}\cite{Schlichtkrull2018}. Their proposal, which is somewhat similar to Defferard's \emph{et al.}, introduce a set of simplifications into a framework of spectral graph convolutions to make them train significantly faster and achieve state-of-the-art levels of accuracy across various classification tasks \cite{Defferrard2016}.

The goal of such models is to learn features on a graph $G = (N, E, Y)$ by taking as input a feature matrix $X$ ($N \times F$ with a feature vector $f_n$ for each node $n$) and a description of the graph structure in the shape of an adjacency matrix $A$ (computed from the set of edges $E$ in the graph). The output is another feature matrix $Z$ ($N \times F'$ with node-level feature vectors $f'_n$ with a predefined number of output features $F'$).

Each \ac{GCN} layer $H^(l)$ in a network with $L$ layers can be expressed as a non-linear function $H^{(l+1)} = f(H^{(l)}, A)$. The first layer takes the input feature matrix ($H^{(0)} = X$) and the final layer generates the output node-level feature matrix ($Z = H^{(L)}$). Each intermediate layer generates a node-level feature matrix $Z^{(l)}$ which is fed to the next layer. In the case of Kipf and Welling \cite{Kipf2016}, the graph-convolution layer $f(H^{(l), A)}$ is defined, in the most basic instantiation, as $\sigma(AH^{(l)}W^{(l)})$, where $\sigma$ is an activation function of choice and $W^{(l)}$ is the weight matrix for the $l$ layer.

This basic framework was heavily extended to overcome two limitations: (1) unless there are explicitly defined self-loops in the graph, the multiplication of $A$ only sums up the feature vectors of all the neighboring nodes but not the node itself, and (2) since $A$ is not normalized by default, the multiplication of $A$ has a huge impact on the scale of the feature vectors. Overcoming those two limitations is crucial to improve the model's convergence.

In order to fix those two limitations, they first enforced self-loops in the graph by adding the identity matrix to $A$ so the new adjacency matrix is $\hat{A} = A + I$. Secondly, they normalized that adjacency matrix in a row-like fashion by leveraging a symmetric normalization with the diagonal node degree matrix $\hat{D}$ of $\hat{A}$. Those two improvements combined form the layer propagation rule proposed by Kipf and Welling \cite{Kipf2016}: $f(H^{(l)}, A) = \sigma(\hat{D}^{-\frac{1}{2}}\hat{A}\hat{D}^{-\frac{1}{2}}H^{(l)}W^{(l)})$. This is the \emph{GCNConv} operator that we used to build our \ac{GNN}.

However, it is important to remark again that this model produces a feature matrix with node-level feature vectors yet our problem needs to classify the whole graph either as stable or slippery. To produce such binary graph-level classification output we need to introduce pooling operations to reduce the amount of nodes in the graph and/or fully connected layers to perform high-level reasoning.

% \begin{table}[!hbt]
%     \centering
%     \caption{Proposed \ac{GCN}-based architecture with five convolutional and two fully connected layers.}
%     \begin{tabular}{c|c|c|c}
%         \hline
%         \textbf{ID} & \textbf{Layer} & \textbf{In Shape} & \textbf{Out Shape}\\
%         \hline
%         conv1 & GCNConv & $B \times 3 \cdot 24 \times 3$ & $B \times 3 \cdot 24 \times 8$\\
%         relu1 & ReLU & $B \times 3 \cdot 24 \times 3$ & $B \times 3 \cdot 24 \times 8$\\
%         conv2 & GCNConv & $B \times 3 \cdot 24 \times 8$ & $B \times 3 \cdot 24 \times 8$\\
%         relu2 & ReLU & $B \times 3 \cdot 24 \times 8$ & $B \times 3 \cdot 24 \times 8$\\
%         conv3 & GCNConv & $B \times 3 \cdot 24 \times 8$ & $B \times 3 \cdot 24 \times 16$\\
%         relu3 & ReLU & $B \times 3 \cdot 24 \times 8$ & $B \times 3 \cdot 24 \times 16$\\
%         conv4 & GCNConv & $B \times 3 \cdot 24 \times 16$ & $B \times 3 \cdot 24 \times 16$\\
%         relu4 & ReLU & $B \times 3 \cdot 24 \times 16$ & $B \times 3 \cdot 24 \times 16$\\
%         conv5 & GCNConv & $B \times 3 \cdot 24 \times 16$ & $B \times 3 \cdot 24 \times 32$\\
%         relu5 & ReLU & $B \times 3 \cdot 24 \times 16$ & $B \times 3 \cdot 24 \times 32$\\
%         fc1 & Fully Connected & $B \times 3 \cdot 24 \cdot 32$ & $B \times 128$\\
%         fc1 & Fully Connected & $B \times 128$ & $B \times 2$\\
%         \hline
%     \end{tabular}
% \end{table}

\section{Experimentation}
\label{sec:experimentation}

We conducted several experiments in order to validate our approach. In this section we describe the dataset we used to carry out such experiments. In addition, we provide all the details of our methodology to ensure the reproducibility of our procedures. At last, we discuss all the experiments that led us to the architecture described in the previous section.

\subsection{Dataset}
\label{sec:dataset}

\begin{figure}[!htb]
	\centering
	\includegraphics[width=0.95\linewidth]{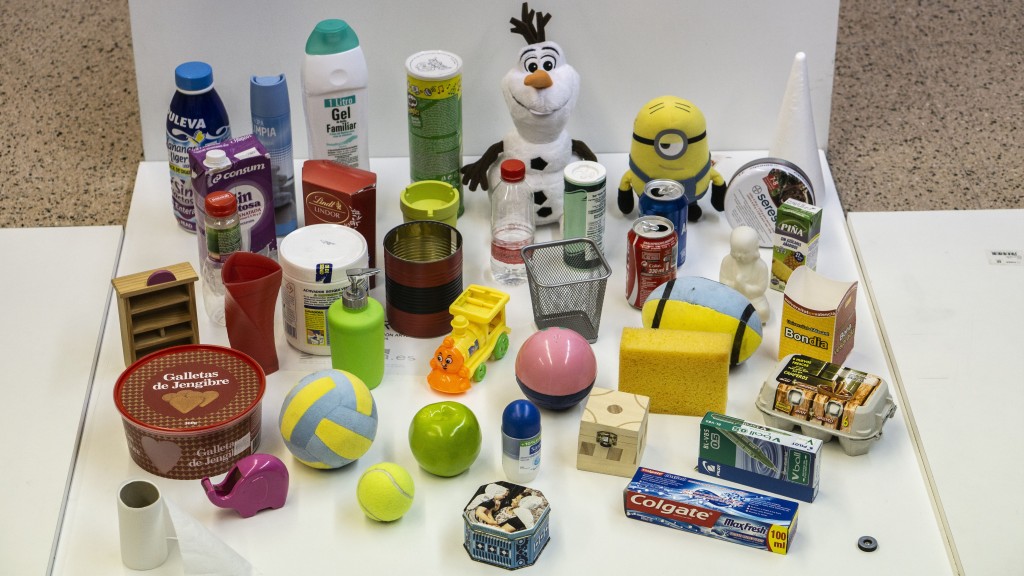}
	\caption{The original training set of $41$ objects.}
	\label{fig:dataset_train}
\end{figure}

\begin{figure}[!htb]
	\centering
	\includegraphics[width=0.95\linewidth]{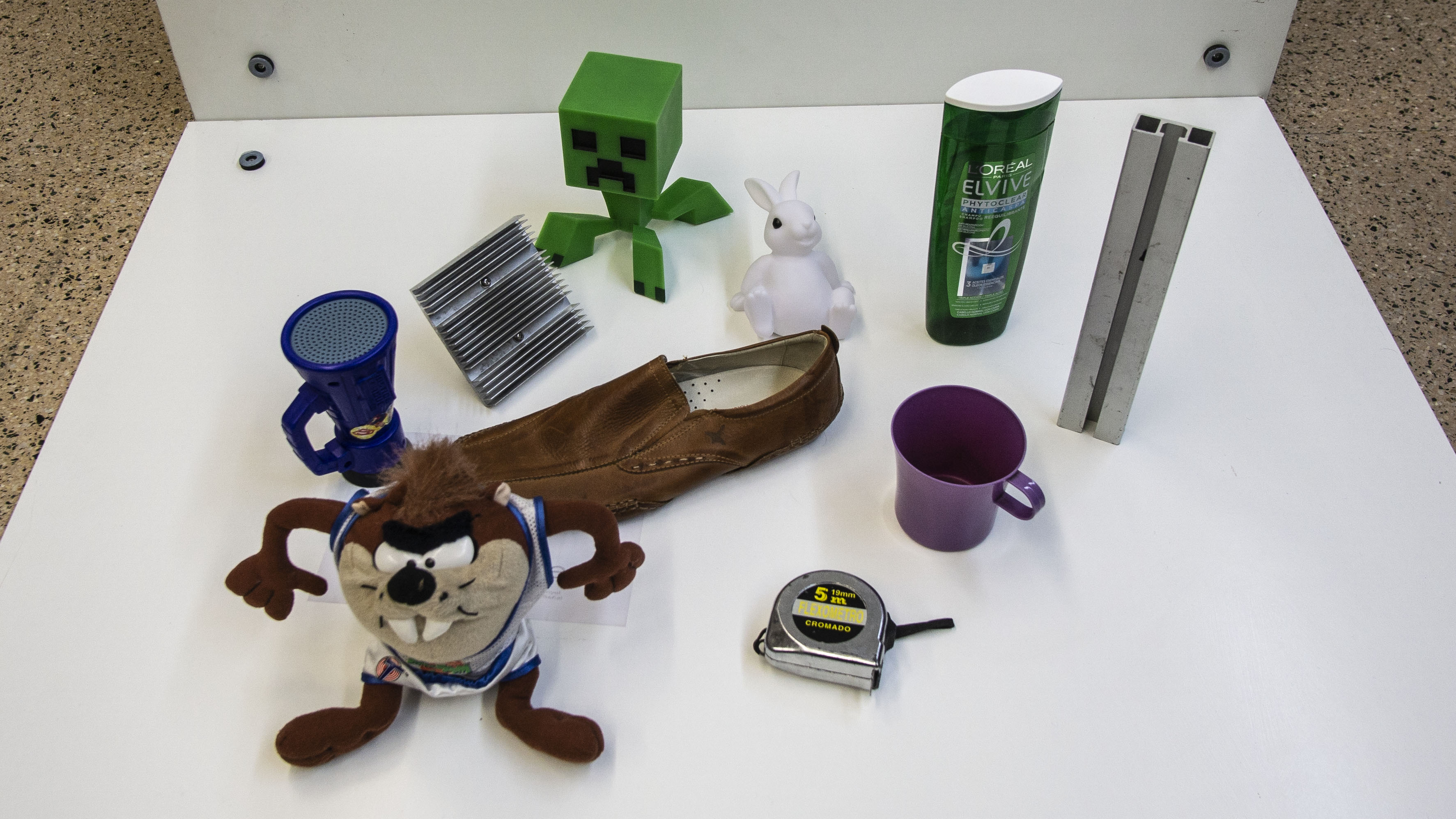}
	\caption{The newly captured test set of $10$ objects.}
	\label{fig:dataset_test}
\end{figure}

The dataset used in our experiments was first introduced in \cite{Zapata2018} as the \emph{BioTacSp Images} dataset. It contains grasp samples performed over $41$ objects with different geometries (i.e. cylinders, spheres, boxes), materials (i.e. wood, plastic, aluminum), stiffness levels (i.e. solid, soft) as well as sizes and weights. Those objects are shown in Figure \ref{fig:dataset_train}. For this work, added $10$ new objects with similar materials but different geometries and stiffness levels (see Figure \ref{fig:dataset_test}). The original $41$ were left for the training set whilst the new ones were separated into a test set. Both sets, training and test, were recorded following these steps:

\begin{enumerate}
	\item \textbf{Grasp the test object:} the hand performed a three-fingered grasp that contacted the object with each of the fingers equipped with a tactile sensor.	
	\item \textbf{Read the sensors:} a single reading was recorded then from each of the sensors at the same time.
	\item \textbf{Lift the object:} the hand was raised in order to lift the object and check the outcome.
	\item \textbf{Label the trial:} the previously recorded tactile readings were labeled according to the outcome of the lifting with two classes (stable, i.e., it is completely static, or slip, i.e., either fell from the hand or it moves within it).
\end{enumerate}

\begin{figure}[!htb]
	\centering
    \includegraphics[width=0.32\linewidth]{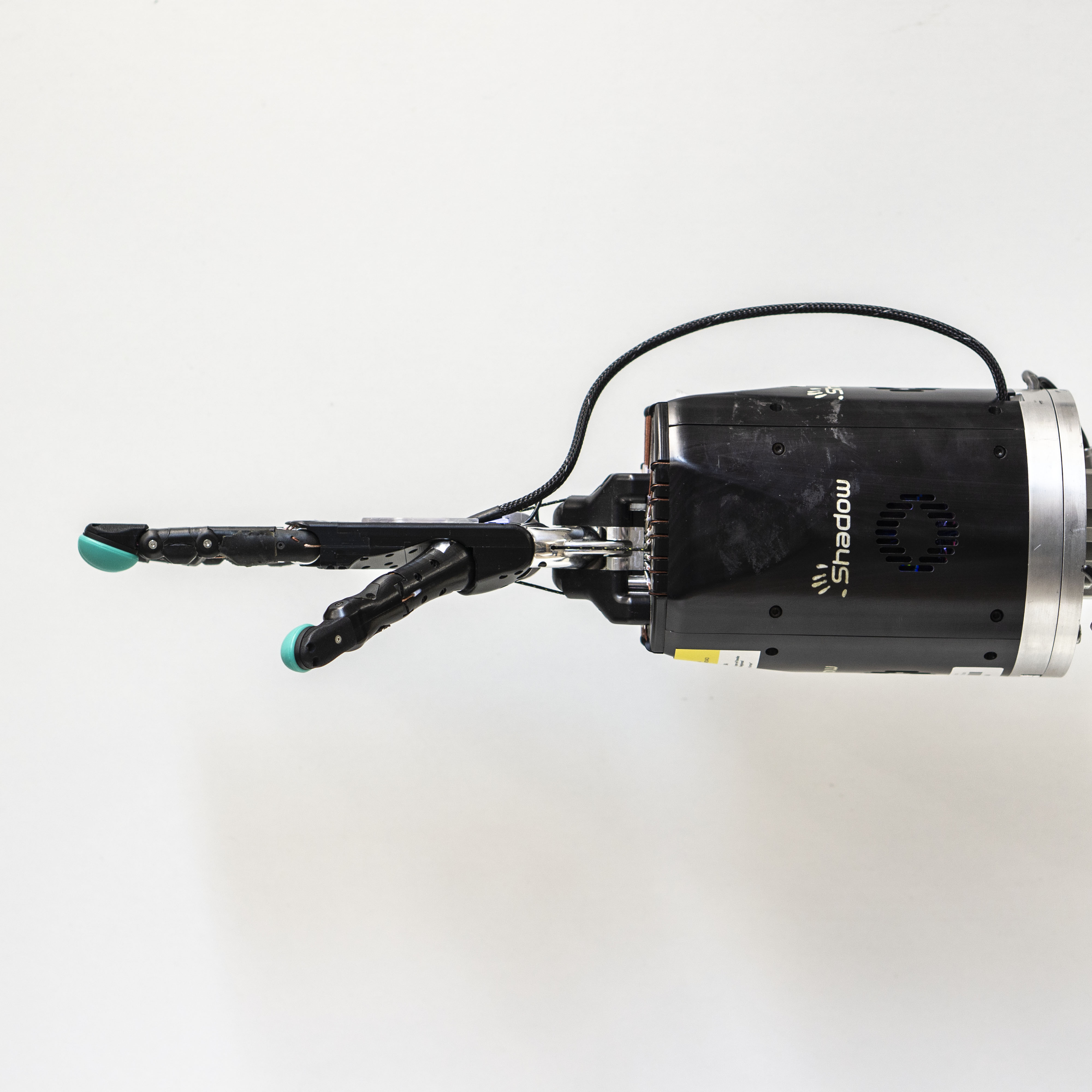}
    \includegraphics[width=0.32\linewidth]{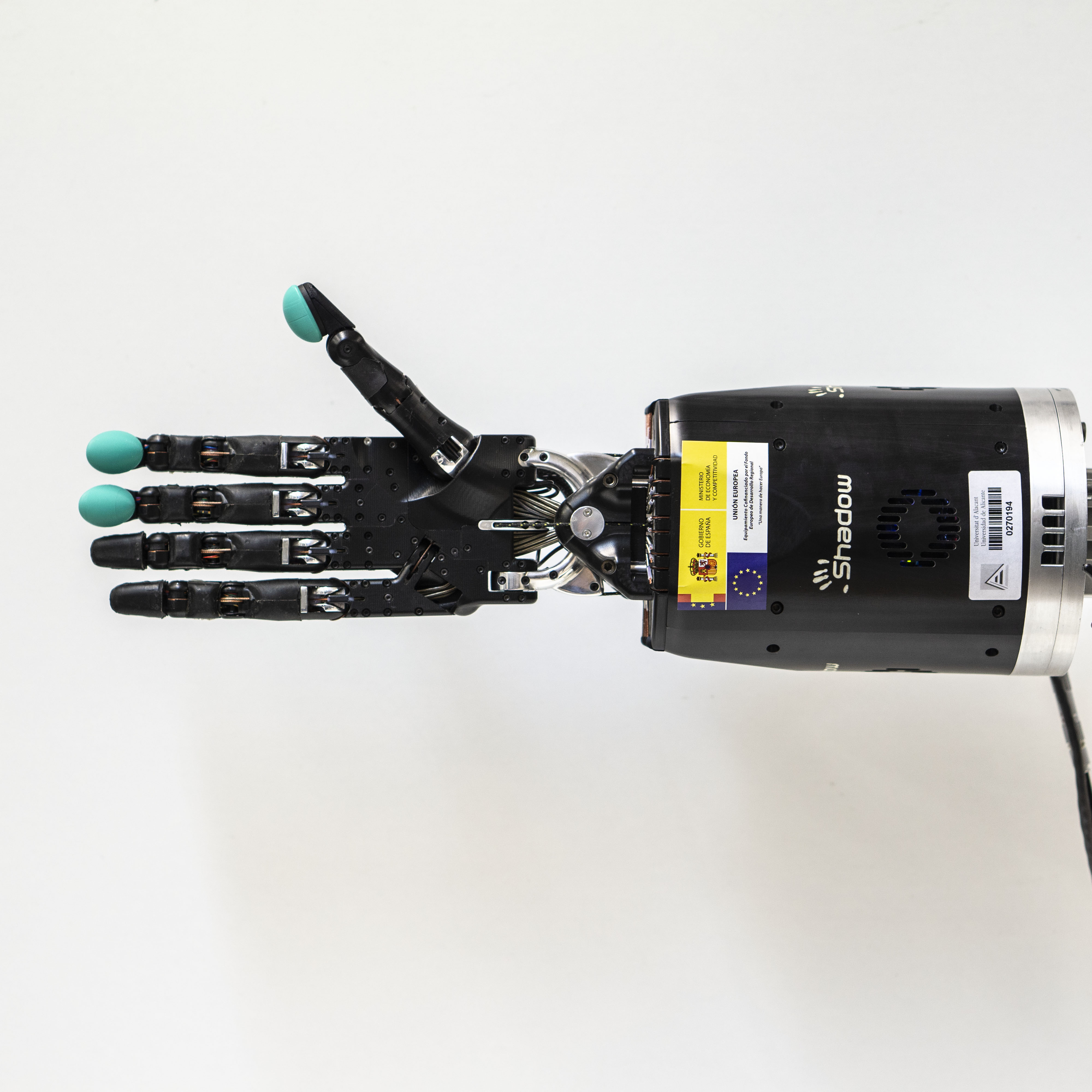}
    \includegraphics[width=0.32\linewidth]{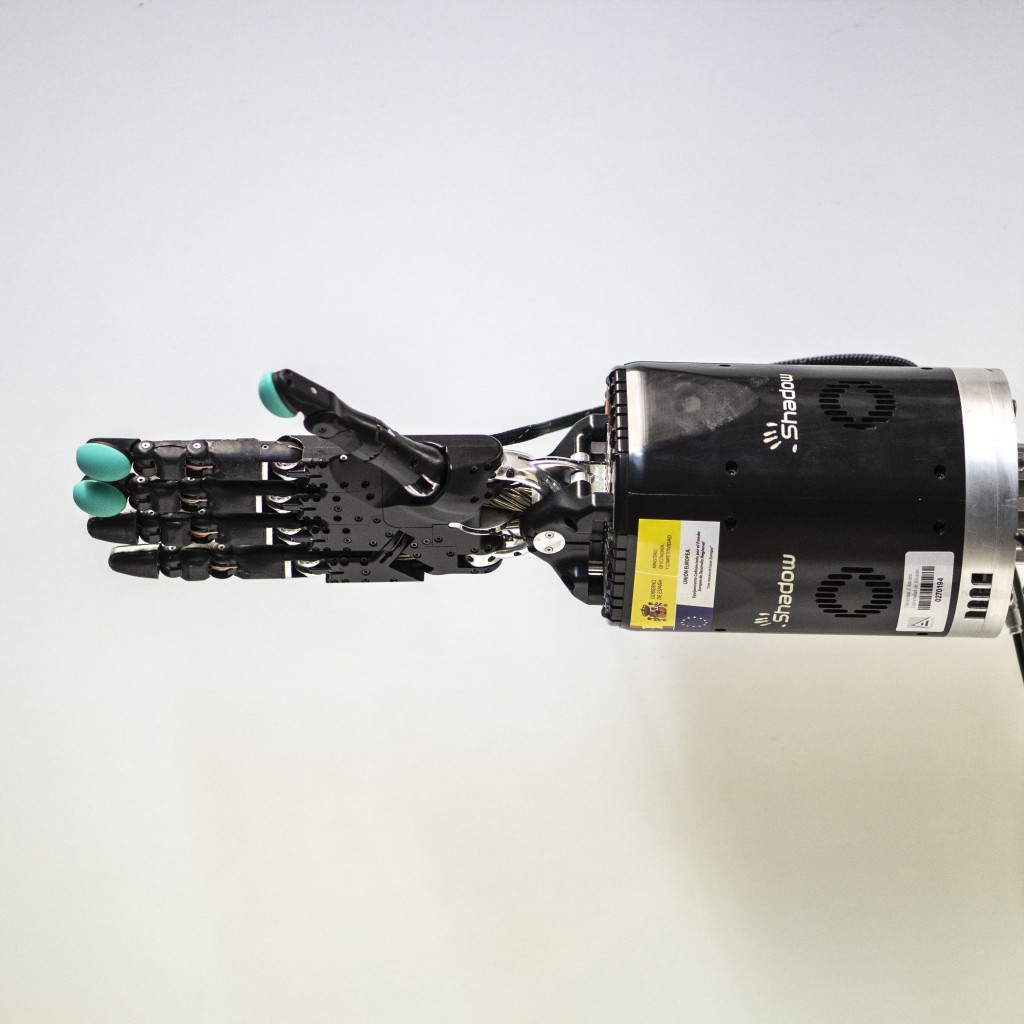}\\
    \smallskip
    \includegraphics[width=0.32\linewidth]{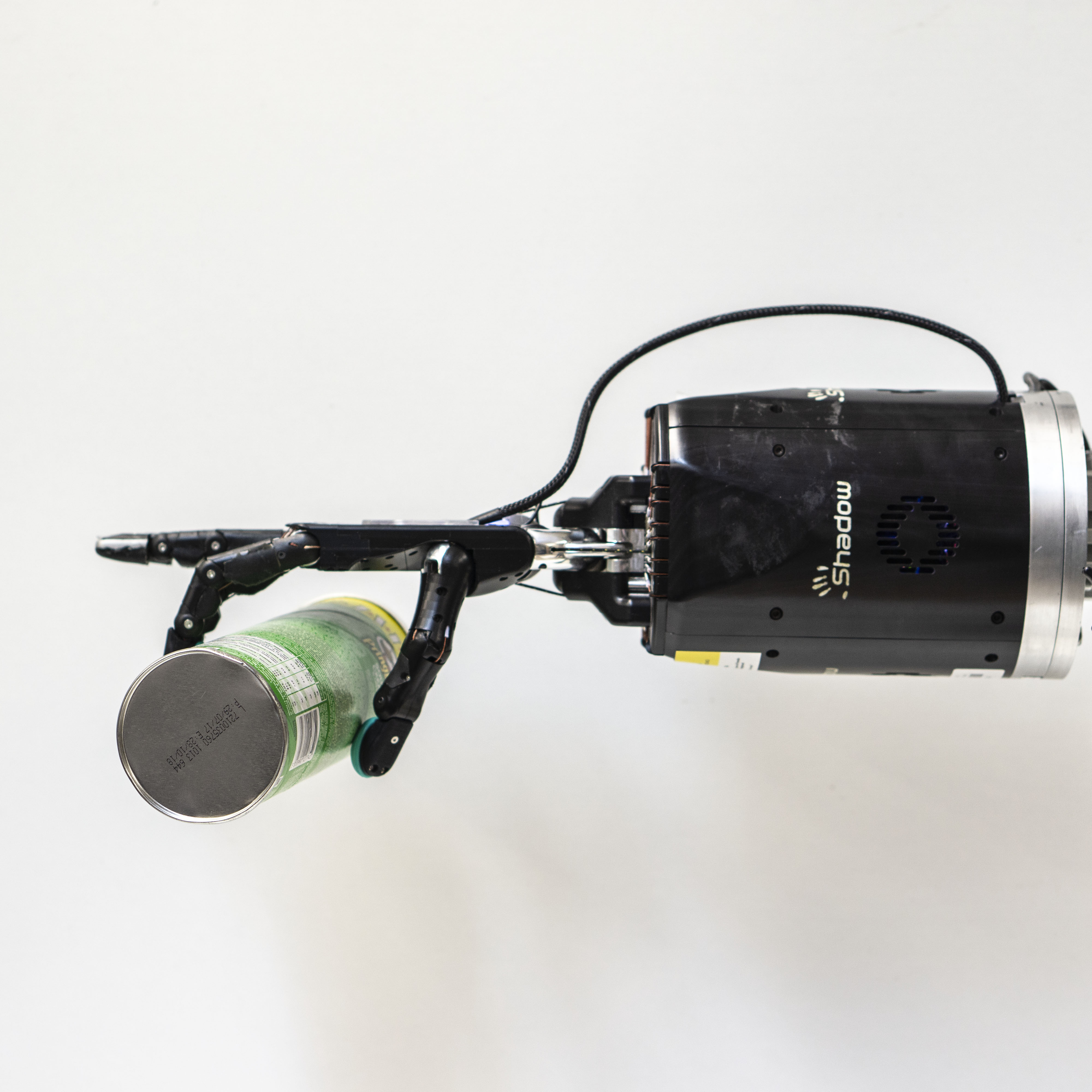}
    \includegraphics[width=0.32\linewidth]{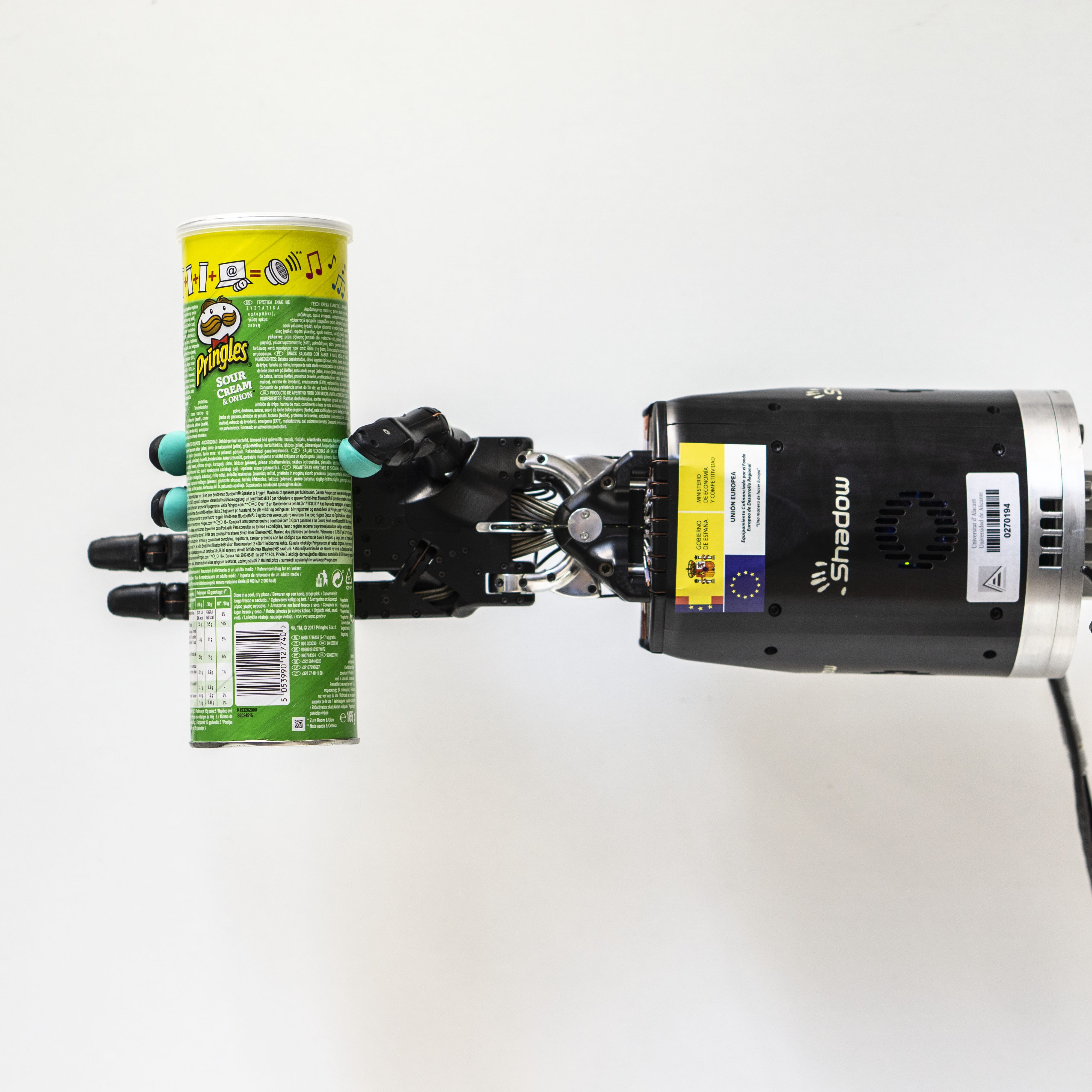}
    \includegraphics[width=0.32\linewidth]{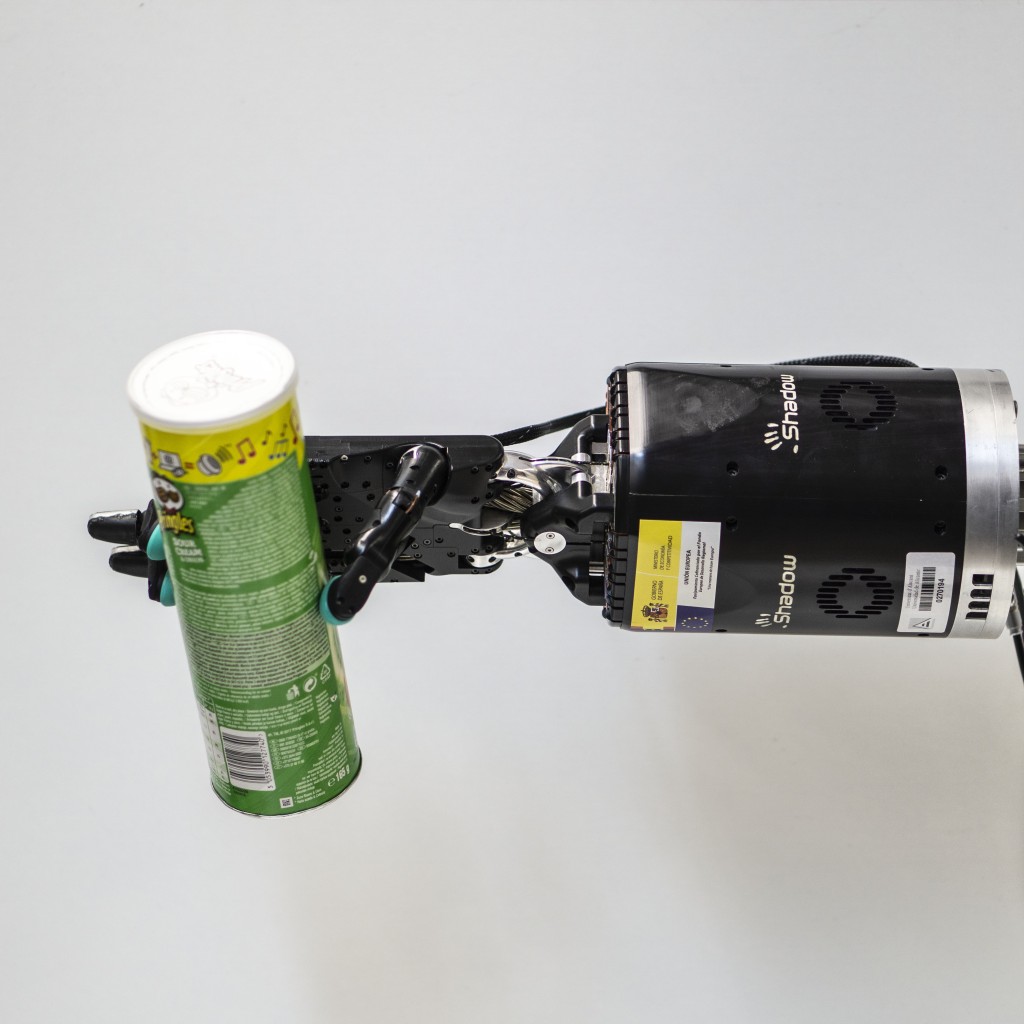}
	\caption{(Top row) Samples of the three hand configurations in the dataset: (from left to right) \emph{palm down}, \emph{palm side}, and \emph{palm 45}. (Bottom row) The same configurations but grasping an object.}
	\label{fig:dataset_grasps}
\end{figure}

There are two hand configurations in the original dataset: \textit{palm down} grasps were performed pointing the palm of the hand downwards while \textit{palm side} grasps were recorded pointing it to one side, with the thumb upwards. In this work, we have added a new configuration: \emph{palm 45} which is in between the other two configurations at an angle of $45$ degrees. Figure \ref{fig:dataset_grasps} shows the aforementioned hand configurations.

Table \ref{table:datasets} provides a quantitative summary of the extended dataset for both splits and all configurations.

\begin{table}[!htb]
	\renewcommand{\arraystretch}{1.3}
	\caption{Summary of the extended BioTacSp dataset which was used in this work to validate our graph-based architecture.}
	\label{table:datasets}
	\centering
	\begin{tabular}{lcccc}
        \hline
        & \multicolumn{2}{c}{\textbf{Training Set}} & \multicolumn{2}{c}{\textbf{Test Set}}\\
        \hline
        \textbf{Configuration} & \textbf{Stable} & \textbf{Slippery}  & \textbf{Stable} & \textbf{Slippery} \\
        \hline
        Palm Down & 667 & 609 & 153 & 163 \\
        Palm Side & 603 & 670 & 157 & 165 \\
        Palm 45 & 1058 & 1075 & 250 & 261 \\
        \hline
        All & 2328 & 2354 & 560 & 589 \\
        \hline           
	\end{tabular}
\end{table}

To the best of our knowledge, there is only one previous work that released a dataset of tactile recordings for the task of grasp stability detection, which is the BiGS dataset \cite{Chebotar2016bigs}. In their work, Chebotar \emph{et al.} recorded 2000 grasps on three standing objects (a cylindrically-shaped box of wipes, a cubically-shaped box of candy and a ball) using a Barret three-fingered hand, which was equipped with three BioTac tactile sensors. Our work extends the BioTac SP
dataset firstly introduced in \cite{Zapata2018}, counting with more than 4000 training grasps and 1000 test grasps with three BioTac SP tactile sensors recorded using 51 objects and various orientations, both for the objects and the hand. The dataset is freely available at GitHub \footnote{\url{https://github.com/3dperceptionlab/biotacsp-stability-set-v2}}.

\subsection{Experimental Setup}

All experiments were run on a computer with an i7-8700 CPU \@ $3.20$ GHz (6 cores / 12 threads) with an Z370 chipset motherboard, 16 GiB DDR4 RAM \@ $2400$ MHz CL15, a Samsung SSD 860 EVO 250 GiB, and an NVIDIA Titan X Maxwell (12 GiB) GPU. Everything was implement in Python $3.6$, PyTorch $0.4.1$, PyTorch Geometric $0.3.1$, CUDA $10.0$ (with driver version $410.73$).

For most experiments, we report accuracy as our main metric to iterate and draw conclusions over training and validation sets. For the test set, we report four different metrics: accuracy, precision, recall, and F1-score (the harmonic mean of precision and recall). To ensure generalization and give an accurate (and statistically correct) estimate of our prediction model performance we employ $k$-fold cross validation with $k=5$. All reported results are the average of $10$ rounds of $5$-fold cross validation. For each cross-validation split, we train our models for $512$ epochs using the ADAM optimizer. The hyperparameters were chosen empirically as follows: $0.01$ learning rate and $5e^{-4}$ weight decay.

The whole source code and dataset for this work can be downloaded from the corresponding GitHub repository\footnote{\url{https://github.com/3dperceptionlab/tactile-gcn}}.

\subsection{Network Depth and Width}

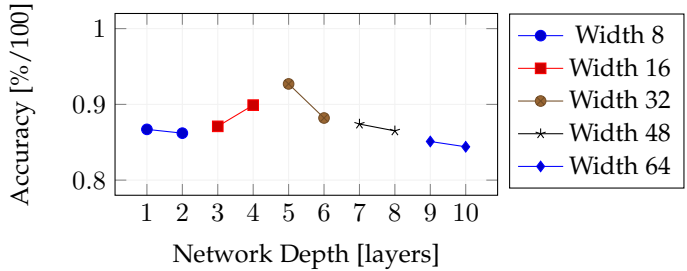
\begin{figure}[!htb]
\begin{tikzpicture}
\begin{axis}[
    width=0.75\linewidth,
    height=4cm,
    title={Validation Accuracy vs Network Depth (and Width)},
    xlabel={Network Depth [layers]},
    ylabel={Accuracy [\%/100]},
    xmin=1, xmax=10,
    ymin=0.8, ymax=1,
    xtick={1, 2, 3, 4, 5, 6, 7, 8, 9, 10},
    ytick={0.0, 0.1, 0.2, 0.3, 0.4, 0.5, 0.6, 0.7, 0.8, 0.9, 1.0},
    legend pos=outer north east,
    grid style={line width=.1pt, draw=gray!10},
    ymajorgrids=true,
    xmajorgrids=true,
    enlargelimits=true,
]

  \addplot
    coordinates {
      (1, 0.867)(2, 0.862)
    };
  \addplot
    coordinates {
      (3, 0.871)(4, 0.899)
    };
  \addplot
    coordinates {
      (5, 0.927)(6, 0.882)
    };
  \addplot
    coordinates {
      (7, 0.874)(8, 0.865)
    };
  \addplot
    coordinates {
      (9, 0.851)(10, 0.844)
    };
  \legend{Width 8, Width 16, Width 32, Width 48, Width 64}
\end{axis}
\end{tikzpicture}
  \caption{Results of network depth and width study.}
  \label{fig:experiments_width_depth}
\end{figure}

In these experiments, we investigate the impact of network depth (convolution layers) and width (amount of features per layer). To that end, we have tested ten different models ranging from one to ten \emph{GCNConv} layers with increasing number of features ($8$, $16$, $32$, $48$, $64$). ReLU activations were used after each convolutional layer. Two fully connected layers were also placed at the end of the network (with $128$ and $2$ output features respectively) to produce the classification
result. We made use of the manually defined graph connections ($k=0$). Figure \ref{fig:experiments_width_depth} shows the results of this set of experiments.

As we can observe, there is a dependency on both width and depth. Shallow networks tend to perform better than their deep counterparts. However, we can find a sweet spot on the architecture with $5$ layers and $32$ features ($8-8-16-16-32$). Shallower networks are not able to fully capture our problem while deeper ones tend to overfit our training data. Consequently, we will proceed with that network.

\subsection{Graph Connectivity}

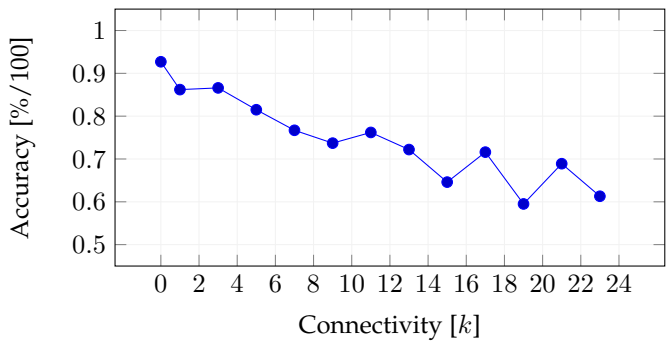
\begin{figure}[!htb]
\begin{tikzpicture}
\begin{axis}[
    width=\linewidth,
    height=5cm,
    title={Validation Accuracy vs Connectivity},
    xlabel={Connectivity [$k$]},
    ylabel={Accuracy [\%/100]},
    xmin=0, xmax=24,
    ymin=0.50, ymax=1,
    xtick={0, 2, 4, 6, 8, 10, 12, 14, 16, 18, 20, 22, 24},
    ytick={0.0, 0.1, 0.2, 0.3, 0.4, 0.5, 0.6, 0.7, 0.8, 0.9, 1.0},
    grid style={line width=.1pt, draw=gray!10},
    ymajorgrids=true,
    xmajorgrids=true,
    enlargelimits=true,
]

  \addplot
    coordinates {
      (0, 0.927)(1, 0.862)(3, 0.866)(5, 0.815)(7, 0.767)(9, 0.737)(11, 0.762)(13, 0.722)(15, 0.646)(17, 0.716)(19, 0.595)(21, 0.689)(23, 0.613)
    };
\end{axis}
\end{tikzpicture}
  \caption{Performance of the network according to the connectivity of the graph.}
  \label{fig:experiments_connectivity}
\end{figure}

For the connectivity experiments we took the previous best network and investigated the effect of graph connectivity. We experimented with manually specified edges ($k=0$) and the \ac{k-NN} strategy with $k = [1, 23]$. As shown in Figure \ref{fig:experiments_connectivity}, the performance of the network degraded as the connectivity of the graph increased in each experiment. Using the \ac{k-NN} strategy, smaller $k$ values achieved greater performance in terms of validation accuracy. However,
none of them improved the performance ($92.7\%$) yielded by the network trained with the graph created using the manual connectivity ($k = 0$).

In the manually created graph there are electrodes connected by an edge to just one other electrode, some others are connected up to four neighbors and the electrode in the center (24th electrode) is connected to six other points. As a result, there are different degrees of connectivity within the graph that could have given some insight to the network about the importance of each node in order to better learn the problem.

\subsection{Generalization Tests}

In order to prove the generalization capabilities of our system, we trained our best network ($8-8-16-16-32$ with $k=0$) with our whole training set and evaluated it on the various test sets (palm down, palm side and palm 45). All results are reported in Table \ref{table:generalization_tests}.

\begin{table}[!htb]
  \centering
    \caption{Results of generalization experiments on the testing splits.}
    \label{table:generalization_tests}
    \begin{tabular}{r|cccc}
        \hline
        \textbf{Test Set} & \textbf{Accuracy} & \textbf{Precision} & \textbf{Recall} & \textbf{F1}\\
        \hline
        Down & 0.741 & 0.741 & 0.751 & 0.745\\
        45 & 0.774 & 0.774 & 0.783 & 0.778\\
        Side & 0.751 & 0.785 & 0.709 & 0.745\\
        \hline
    \end{tabular}
\end{table}

There is a significant drop in accuracy when dealing with completely unknown objects. Recall that the test set consists of new objects with different geometries and stiffness levels so they are substantially different from the training set. Taking all of this into account, and despite the difficulty of the testing set, we can expect gains from applying regularization and augmentation strategies to increase performance on data whose distribution is not that similar to the training set.

\section{Conclusion}
\label{sec:conclusion}

Tactile sensors provide useful information for robotic manipulation tasks like predicting grasp stability. Prior works in the literature tend to compute hand-engineered features that are later used for training a machine learning model. A recent trend process them as images, so deep learning techniques like \acsp{CNN} can calculate relevant characteristics that lets the system distinguish a slippery grasp from a stable one. Inspired by this methodology, we propose in this work a novel approach to tactile data interpretation: we build a graph with the sensor's taxels because this structure keeps more accurately the spatial distribution and the local connectivity of these sensing points. The goodness of these properties and the tactile graph for grasp stability prediction were tested in experimentation.

We used three BioTac SP tactile sensors mounted in the tip of the index, middle and thumb of a Shadow Dexterous hand. In order to predict grasp stability using these graph representations of the tactile sensors, we trained a \ac{GCN} with a custom dataset which was captured with more than 50 objects and 3 hand orientations. The robustness of the proposed system was checked by testing the system with novel orientations and objects. In average, the \ac{GCN} yielded a 92.7\% validation accuracy on the prediction of grasp stability with novel objects or orientations.

\section{Limitations and Future Works}
\label{sec:limitations}

Given the obtained results, graph representations of tactile readings can be successfully used for learning the task of grasp stability prediction. Nevertheless, there are some drawbacks linked to their used. The first limitation of this proposal is the problem of defining the graph connectivity. We had to find a way of defining the location of the taxels as well as their connections in order to define the graph. In the case of using the tactile readings directly, none of this is necessary.

Moreover, \ac{GCN} showed to be data hungry models for learning. In a previous work \cite{Zapata2018}, the authors obtained higher performance rates with fewer data samples for training a \ac{CNN}. For this work, it was necessary to capture more data in order to achieve similar accuracy rates in training. Furthermore, generalization to radically new objects has still a lot of room for improvement by leveraging techniques such as L2 regularization, dropout, or data augmentation
itself.

As a future work, we also plan to decouple the currently unified \ac{GCN} for the three fingers so that each graph is processed by a different network path. Furthermore, we plan to model the noise of each individual taxel and augment each sample on the fly by adding random noise following each taxel's distribution. At last, we plan to extend the architecture to predict grasp stability over temporal sequences by fusing the \ac{GCN} model with \ac{LSTM} networks in a similar way as Conv-\ac{LSTM} models do.

% BRAYAN: Quiza debemos poner el proyecto nuestro de COMMANDIA que va sobre temas relacionados y es el que estamos trabajando ahora... lo preguntare.

\section*{Acknowledgment}

Experiments were made possible by generous hardware donations from NVIDIA (Titan Maxwell). This  work  has  been  funded  by  the  Spanish  Government TIN2016-76515-R and  DPI2015-68087-R grants supported  with  Feder  funds.  This  work  has  also  been  supported  by two grants for PhD studies (FPU15/04516 and BES-2016-07829). In addition, it has been funded by regional projects GV/2018/022 and GRE16-19.

\bibliographystyle{IEEEtran}
\bibliography{references}

\end{document}